\documentclass[10pt,twocolumn,letterpaper]{article}

\usepackage{iccv}
\usepackage{times}
\usepackage{epsfig}
\usepackage{graphicx}
\usepackage{amsmath}
\usepackage{amssymb}
\usepackage{epigraph}

\usepackage{duckuments}
\usepackage[utf8]{inputenc} %
\usepackage[T1]{fontenc}    %
\usepackage{url}            %
\usepackage{booktabs}       %
\usepackage{amsfonts}       %
\usepackage{nicefrac}       %
\usepackage{microtype}      %
\usepackage{epsfig}
\usepackage{graphicx}
\usepackage{amsmath}
\usepackage{amssymb}
\usepackage{booktabs}
\usepackage{wasysym}
\usepackage{xcolor} 
\usepackage{bbm}
\usepackage{multirow}
\usepackage{mathtools}
\usepackage{wrapfig}
\usepackage{cancel}
\usepackage{ulem}
\usepackage[font=scriptsize,labelfont=bf]{caption}
\usepackage{subcaption}
\usepackage{algorithm}
\usepackage{listings}
\usepackage{textpos}  % for superposing a text box

\DeclareMathOperator*{\argmax}{arg\,max}

\usepackage[pagebackref=true,breaklinks=true,letterpaper=true,colorlinks,bookmarks=false]{hyperref}

\iccvfinalcopy % *** Uncomment this line for the final submission

\def\iccvPaperID{8241} % *** Enter the ICCV Paper ID here

\ificcvfinal\fi

\newcommand{\dataname}{FunKPoint}
\newcommand{\tablestyle}[2]{\setlength{\tabcolsep}{#1}\renewcommand{\arraystretch}{#2}\centering\footnotesize}

\begin{document}

%%%%%%%%% TITLE
\title{The Functional Correspondence Problem\vspace{-10pt}}

\author{
Zihang Lai$^*$\hspace{2em}
Senthil Purushwalkam$^*$\hspace{2em}
Abhinav Gupta\vspace{0.5em}\\
Carnegie Mellon University\\}
\maketitle

%%%%%%%%% ABSTRACT
\begin{abstract}
The ability to find correspondences in visual data is the essence of most computer vision tasks. But what are the right correspondences? The task of visual correspondence is well defined for two different images of same object instance. In case of two images of objects belonging to same category, visual correspondence is reasonably well-defined in most cases. But what about correspondence between two objects of completely different category -- e.g., a shoe and a bottle?  Does there exist any correspondence? Inspired by humans' ability to: (a) generalize beyond semantic categories and; (b) infer functional affordances, we introduce the problem of functional correspondences in this paper. Given images of two objects, we ask a simple question: what is the set of correspondences between these two images for a given task? For example, what are the correspondences between a bottle and shoe for the task of pounding or the task of pouring. We introduce a new dataset: FunKPoint that has ground truth correspondences for 10 tasks and 20 object categories. We also introduce a modular task-driven representation for attacking this problem and demonstrate that our learned representation is effective for this task. But most importantly, because our supervision signal is not bound by semantics, we show that our learned representation can generalize better on few-shot classification problem. We hope this paper will inspire our community to think beyond semantics and focus more on cross-category generalization and learning representations for robotics tasks.
\end{abstract}
{\let\thefootnote\relax\footnote{{* Authors contributed equally}}}

\begin{textblock*}{.8\textwidth}[.5,0](0.5\textwidth, -.775\textwidth)
\centering
{\small Website: \url{https://agi-labs.github.io/FuncCorr}}
\end{textblock*}

%%%%%%%%% BODY TEXT
\section{Introduction}
\epigraph{To perceive an affordance is not to classify an object. The fact that a stone is a missile does not imply that it cannot be other things as well. It can be a paperweight, a bookend, a hammer, or a pendulum bob.}
{\textit{James J. Gibson}}

\begin{figure}[t!]
    \centering
    \includegraphics[width=\linewidth]{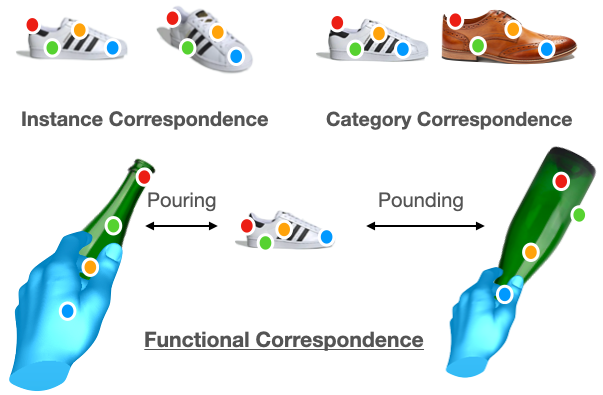}
    \caption{Given a pair of images, functional correspondence establish correspondence between points that are \textit{functionally} the same. In this example, we hold the body of the bottle when pouring but the neck when pounding, therefore we can establish correspondence (bottle body, shoe front) for pouring and correspondence (bottle neck, shoe front) for pounding.}
    \vspace{-1em}
    \label{fig:teaser}
\end{figure}

Computer vision and visual representation learning has been bound by shackles of semantic categories. Our training data is built with semantic categories - ImageNet has 1K categories of breeds of dogs, cats and mushrooms. Our supervision is semantic categories. And our evaluation tasks are semantic -- image classification, object detection, image segmentation and list goes on. So it is not surprising that our approaches are bound by the limits of semantic categories. Our representations are not effective in capturing affordances for robotics tasks. And our representations fail to generalize effectively to new object categories due to focus on learning  intra-class invariances. On the other hand, humans have marvelous ability to think beyond categories. We can use a screwdriver for opening screws but also to clean printer, hammer nails and what-not. Clearly, our current semantically-driven computer vision needs rethinking.  

% \epigraph{Correspondence, Correspondence, Correspondence}{\textit{Takeo Kanade\\} in response to 3 most important problems in vision}

In classical computer vision, semantics did not play such an important role. Instead, correspondence was cited as one of the most important tasks in the field of computer vision. It is also the fundamental goal of visual representation learning -- an embedding space where similar objects/parts/pixels have similar embedding. In an anecdotal conversation about the three most important problems in computer vision, Takeo Kanade stated that they are ``\textit{Correspondence, Correspondence, Correspondence}". Yet, this fundamental task of visual correspondence is ambiguous and ill-defined. What is visual correspondence? Does there exist correspondence between any pair of images? The visual correspondence problem is most well-defined and often studied in context of tracking and multi-view reconstruction where the goal is to create correspondences between two images of same object~\cite{Wang19}. It has also been studied in the context of semantic categories where the goal is to create correspondences between images of object instances from same categories~\cite{Kim17,Novotny17}. But it often stops at cat what are the right correspondences between two seemingly different object categories (for example, a bottle and a shoe)? 

In contrast, we humans can identify correspondences between semantically different objects. We unconsciously use this ability to transfer our object manipulation skills to novel objects in order to efficiently accomplish everyday tasks. Specifically, humans possess three interesting capabilities: (a) the ability to visually infer affordances for objects, (b) the ability to generalize beyond semantic categories and (c) the ability to adapt affordances for different tasks.  In order to facilitate exploration of these capabilities, we introduce the problem of functional correspondence. Given images of two objects, we ask a simple question: for a given task, what would be the set of correspondence between two objects? For example, the correspondences between shoe and bottle for the task of pouring are shown in the figure~\ref{fig:teaser}. The grasp locations are shown by green, storage by orange and pouring spout by red keypoints. On the the other hand, the correspondences between shoe and bottle for the task of pounding (hitting with a force) are quite different and shown in figure~\ref{fig:teaser}. Note that the correspondence between two objects is driven by both 3D shape and physical/material properties.

We also introduce a new dataset called \dataname{} (Section \ref{sec:data}). \dataname{} has ground-truth keypoints labeled for 10 tasks across 20 object categories. We also propose a modular task-driven architecture. More specifically, our modular architecture computes the image representation given an input task. We show our architecture is highly effective in modeling functional correspondences although there is still a significant gap with respect to human performance. But most importantly, in proof-of-concept experiments, we demonstrate the underlying promise of learning functional correspondence. Because our task has functional supervision and there is cross-category supervision, our representation can outperform semantically-learned representations for few-shot learning.
%both robotics tasks such as grasping and few-shot learning respectively.

\subsection{Why Functional Correspondence?}
In this paper, we introduce the problem of functional correspondence. We believe this task forms the core of visual learning because of the following reasons: 

(a) \textbf{Object Affordances and Functional Representations:} Ability to predict object affordances is a cornerstone of human intelligence and a key requirement for robotics tasks. The task of functional correspondence allows us to learn functional representations useful for robotics tasks. But more importantly, beyond predicting primary affordances (screwdriver is used for screwing), humans are really good at predicting secondary affordances (how we can use novel objects to fulfil the task -- e.g. using screwdriver to clean paper jam in printer). Modeling functional correspondences across different object categories should help in predicting novel use of objects.

(b) \textbf{Generalization Beyond Semantic Categories:} Unlike other vision tasks such as object classification/detection or even learning 3D from image collections, this task cuts across object semantics. It attempts to model commonalities across different categories of object and hence open up the possibility of generalization beyond semantic categories.

(c) \textbf{Task-Driven Representation:} Finally, the ground-truth is conditioned on the task itself, the correspondences between pair of objects depends on how you envision using these objects. This allows us to formulate a task-driven representation (unlike current existing task-agnostic ConvNet representations).

\section{Related Work}
\noindent \textbf{Correspondences:}
The correspondence problem has always been a focus of the computer vision community, and many sub-problems have been proposed with solutions offered. 
The classical correspondence problem establish correspondence between different views of the same object. 
Such correspondence is crucial for multi view geometry based algorithms and are typically solved by matching local descriptors of interesting points~\cite{bay2006surf, berg2005shape,harris1988combined, lowe1999object, lowe2004distinctive}. 
More recently, researchers looked into category-level correspondence~\cite{Rocco17, Rocco18,ufer2017deep,Kim17,Novotny17}, which does not restrict correspondence to a single instance. Such methods often model correspondence in deep feature space, and relies on simulated transformations for training. 
Because object of the same category usually perform similar actions, our work could also establish correspondence at category level. 
However, we consider any object, regardless of its object class, could correspond if they share parts that have similar functional semantics. 
Thus, our \textit{functional correspondence} could be considered more general in that we also establish cross-category correspondences.

Dense correspondence between pixels across video frames (optical flow) is also studied as a separate problem. Traditionally, the optical flow estimation problem is addressed as an energy minimization problem based on color constancy~\cite{horn1981determining, Brox09, Revaud15, Smith95b}.
Recent optical flow estimation algorithms make use of neural networks~\cite{Ilg17,Bailer17,Jia17,kendall2017end} as models and explores self-supervision as the training method~\cite{liu2019selflow,godard2019digging}. Another line of work focuses on the mid-level optical flow problem~\cite{Wang19,Lai19,jabri2020space} where  consistency between the regions around the pixels is also considered. 
Such approaches often leverage the spatial temporal coherence nature of videos to provides a natural supervision signal. 
However, because the main training loss is usually a photometric loss, the learned correspondence is inevitably local. 
In this work, we try to establish a higher level \textit{functional correspondence}. 
Such correspondence involves a knowledge of object affordances, 
which is still hard to learn from unlabeled raw videos.

\noindent \textbf{Functional Representations and Affordances:} The core idea of affordances was introduced by James J. Gibson~\cite{Gibson79}. Gibson described object affordances as ``opportunities for interactions''. Inspired by Gibson's idea of affordances, a long-term goal for robotic perception has been to perform function recognition~\cite{Stark, Rivlin}. Approaches such as \cite{Stark, Binford} used manually-defined rules to predict affordances. However, these approaches were too brittle and failed to generalize.

In recent years, with the advances in 3D scene understanding and with the large-scale availability of interaction data the idea of affordances has been revisited as well~\cite{Gupta11, Grabner11, Fouhey12, Zhao13, Xie13}. Approaches such as ~\cite{Gupta11, Zhao13} have attempted to use 3D understanding followed by affordance estimation. More recently, approaches have tried to collect large-scale data for affordance estimation~\cite{BingeWatching} and used ConvNets to predict affordances in the scene~\cite{FouheyDirect, Wang21}. AffordanceNet~\cite{AffordanceNet18} simultaneously localizes multiple objects and predicts pixel-wise affordances by training on a large-scale dataset with affordance labels. In contrast, our approach focuses on affordances as a vehicle to target generalization beyond semantic categories and learn task-driven representations. More specifically, we target using primary and secondary object affordances to learn visual correspondences across different object categories. Our work is also closely related to some recent work in robotics which focuses on extraction of keypoints for robotics tasks~\cite{manuelli18, manuelli19}. However, in most of these scenarios, the goal is to learn to predict dense keypoints/correspondences across two objects of same categories. In this work, we focus on the more general problem of how to do task-driven functional correspondences across multiple object categories.

\noindent \textbf{Task-Driven Representations and Modular Networks:}
Classification models in deep learning have largely been trained as discriminative models\cite{krizhevsky2012imagenet,simonyan2014very,he2016deep}. Recently, energy based models\cite{lecun2006tutorial} have gained popularity and demonstrated success on image classification\cite{grathwohl2019your}, continual learning\cite{du2019implicit}, compositional zero-shot learning\cite{purushwalkam2019task,wang2019tafe} and generative modeling of text\cite{bakhtin2021residual}. In \cite{purushwalkam2019task,wang2019tafe}, the key idea is to construct a task-dependent (or label-dependent) neural network for classifying whether an image belongs to the considered label. 
In \cite{purushwalkam2019task}, this compatibility of an image $x$ to a label $y$ is computed using a sequence of neural network modules which are reweighted using a function of the considered label $y$. The modular architecture proposed in \cite{purushwalkam2019task} allows sharing of learned filters across different labels which is crucial for domains where the labels are heavily related. These modular neural networks have also demonstrated great success in multi-task reinforcement learning\cite{devin2017learning,yang2020multi} where modules are shared among related tasks to learn policies efficiently. For estimating functional correspondences, we require representations that vary according to the considered task. Therefore, we adopt a similar modular task-driven architecture for learning a task-dependent representation which also allows us to share neural network modules between related tasks. 
\begin{figure*}[t!]
    \centering
    \includegraphics[width=0.9\textwidth]{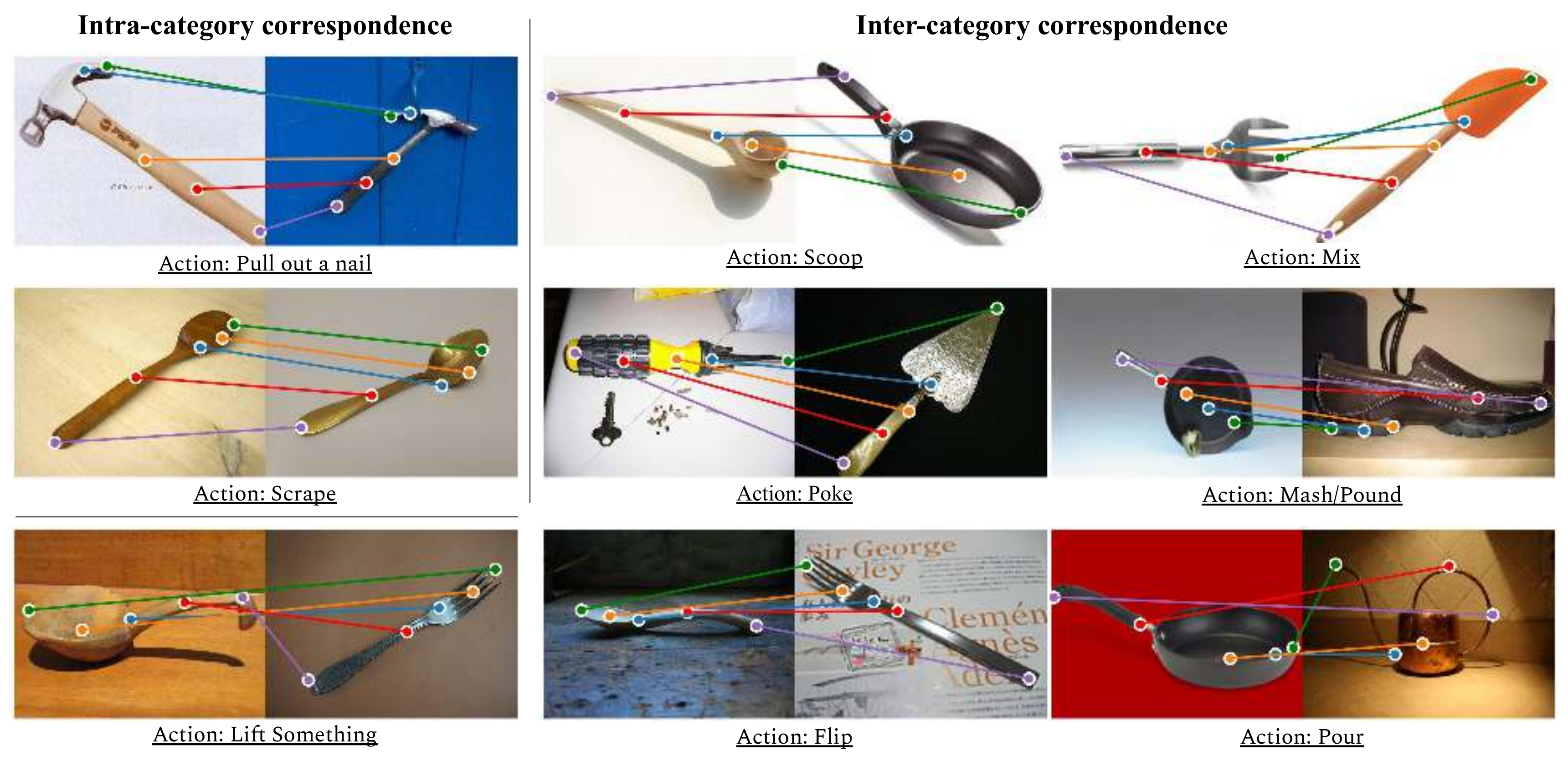}
    \vspace{-1em}
    \caption{\textbf{The \dataname{} Dataset:} Here we present examples from the proposed dataset. For each image and associated task, we collect human annotations for 5 keypoints. Associating keypoints between images provides us with numerous intra-category and inter-category functional correspondences. 
}
    \vspace{-1em}
    \label{fig:dataset}
\end{figure*}

\section{The \dataname{} Dataset}
\label{sec:data}
To explore the study of functional correspondences, we present a novel dataset: \dataname{} (short for Functional KeyPoints). \dataname{} consists of 2K objects covering 20 object categories. In order to learn and evaluate functional correspondences between pairs of images, we require dense human annotations of such correspondences. However, such an approach is unscalable due to the quadratic number of image pairs and pixels.  Instead, we first identify 5 semantically meaningful points that are essential for each task. For each task, we then collect annotations for the 5 keypoints for each relevant object image. Figure~\ref{fig:dataset} shows examples from the dataset. Note that a single image could be  labeled differently for each task. In total, around 24K such labeled keypoints are obtained. Any two objects that can be used to perform an action are then used to establish a correspondence relationship (w.r.t. that action). This correspondence between two images, conditioned on a specific action, is referred to as a \textit{Functional Correspondence}. For example, in the top left figure of Fig.~\ref{fig:dataset}, both hammers can be used to pull out a nail, so a \textit{functional correspondence} relationship (consists of 5 pairs of corresponding points) could be established between the two objects. Similarly, both the spoon and the frying pan (Fig.~\ref{fig:dataset} top-middle) can be used to scoop things, so we can also generate a \textit{functional correspondence} relationship between them. \\

\noindent
\textbf{Data Collection}~~ 
First, we curate an action vocabulary consisting 10 common tasks (or actions).
Our action vocabulary is inspired from the TaskGrasp~\cite{murali2020taskgrasp} dataset, which focuses on task-dependent robot grasps.  Therefore, the 10 actions in our vocabulary are not only common, but also useful as a benchmark in robotics. For each action, we identify 5 object categories that can be used to perform that task. Note that many object categories can be relevant for multiple tasks. This allows us to generate different correspondence for the same objects under the condition of performing different tasks. For example, the object category \textit{frying pan} has 2 possible actions (among others): \textit{Scoop} and \textit{Mash/Pound}. The rim of the pan is a functional keypoint that is important for scooping, but for pounding, the bottom of the pan becomes the relevant functional keypoint. See Table~\ref{tab:object} for the list of 20 objects and their associated tasks. 
\begin{table}[h]
\centering
\tablestyle{8pt}{1.05}

\caption{\textbf{Object categories corresponding to 10 action classes used in \dataname{}:}  
}
\vspace{-1em}
\label{tab:object}
\begin{tabular}{l r}
\toprule
\textbf{Action}          & \textbf{Objects} \\
\midrule
Pour   & bottle, frying pan, watering can, cup, dustpan \\
Scoop  & spoon, basket, cup, frying pan, shoe \\
Mix & spoon, tablefork, spatula, tongs, whisk \\
Mash/Pound & bottle, frying pan, hammer, ladle, shoe \\
Lift Something & ladle, tablefork, basket, tongs, dustpan \\
Scrape & scraper, tablefork, spatula, trowel, spoon \\
Poke & scraper, watering can, screwdriver, trowel, scissors \\
Brush/Dust & whisk, scrub brush, toothbrush, scraper, spoon \\
Pull out a nail & hammer, ladle, scissors, frying pan, tablefork \\
Flip & spoon, tablefork, spatula, ladle, tongs \\
\bottomrule
\vspace{-2em}
\end{tabular}

\end{table}

For each of the 20 object categories, we collect 100 images from the ImageNet dataset~\cite{deng2009imagenet}, but supplementing with creative commons images from Google image search to reach 100. Note we manually filter out images that contain multiple object instances, missing parts or occluded parts. 
% Finally, 17 out of 20 classes have at least 100 images and the remaining 3 classes have at least 60 images.

We use Amazon Mechanical Turk to collect human annotations for the keypoints. Each (image, task) pair is labeled with the 5 functional key points as well as a choice of labelling difficulty (between easy, medium or hard). In the interface, we provide a simple definition for each point, current action, and also examples of labeled images. See supplementary material for a visualization of the interface. As explained, each object could be associated with multiple actions (see Supp. for statistics). From the collected data, we create a train split containing 4044 (image, task) pairs and a test split contains 741 (image, task) pairs.

\section{Approach}

\begin{figure*}[t!]
    \centering
    \includegraphics[width=0.85\textwidth]{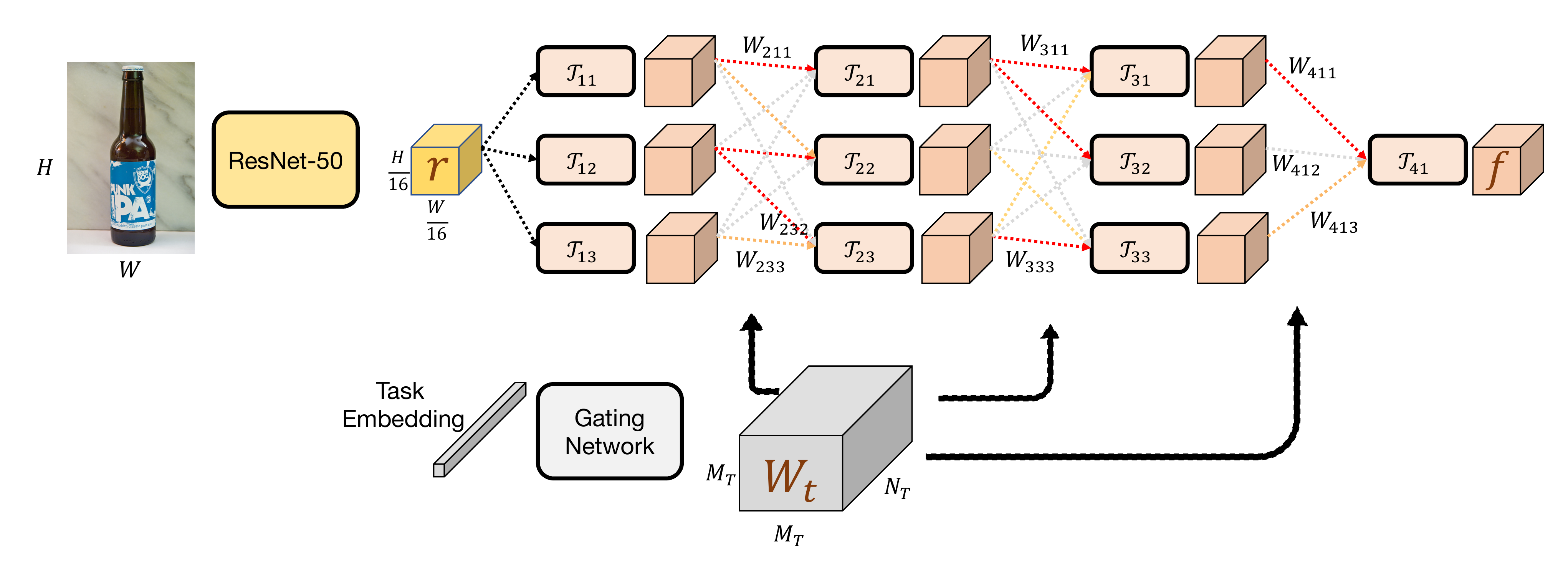}
    \vspace{-1em}
    \caption{\scriptsize \textbf{Approach:} We use a task-driven modular architecture for learning functional representations. We show that the learned representation can be effectively used to identify functional correspondences between objects. Note that we show 3 modules per layer here only for illustration, see supplementary material for the chosen value for this hyperparameter. 
}
\vspace{-1em}
    \label{fig:approach}
\end{figure*}

Estimating \textit{semantic} correspondences has been well studied in the past. Most approaches\cite{Rocco17, Rocco18,ufer2017deep,Kim17,Novotny17} involve learning a pixel or patch level representation which can be used to match corresponding points on similar objects. As we will demonstrate via experiments, for the problem of functional correspondence, such representations are not suitable. We wish to estimate correspondences even across semantically varied objects and second, the correspondences vary according to the task being performed. Therefore, we propose an approach that produces task-driven representations that can be used to find functional correspondences across varied objects. 

First, we formalize the problem setup of functional correspondence. Consider two images $\mathcal{I}$ depicting an object $o$ and $\mathcal{I}'$ depicting an object $o'$ such that both objects can be used to perform task $t$. Given any point $p$ on object $o$, the goal of the functional correspondence problem is to estimate the functionally corresponding location $p'$ on object $o'$. 
However, as described in Sec~\ref{sec:data}, we only have access to correspondences for specific keypoints due to the prohibitive cost of annotation. Therefore, for each task $t\in\mathcal{T}$ that can be performed with object $o$, we have a set of functional keypoints $\{p_{t1}, p_{t2}, ..., p_{tK}\}$. The goal of the functional correspondence problem can then be restated as estimation of the functionally corresponding locations of the keypoints $\{p'_{t1}, p'_{t2}, ..., p'_{tK}\}$ on object $o'$.

Recently, task-driven classifiers have gained popularity for the problem of zero-shot learning\cite{purushwalkam2019task,wang2019tafe}. Taking inspiration from these approaches, we adopt a similar approach to learn a task-driven representation. More formally, we propose a model $\mathcal{F}_\theta$ with parameters $\theta$ which takes as input an image $\mathcal{I}$, a task $t$ and outputs a representation $f=\mathcal{F}_\theta(I,t)$. In order for the representation $f$ to be useful for functional correspondence, we propose to learn the parameters $\theta$ using the dataset presented in Sec~\ref{sec:data}. The goal  is to ensure that the representation $f$ at location  $p$ of an image $\mathcal{I}$ and location $p'$ of an image $\mathcal{I}'$ are identical only when $p,p'$ are functional correspondences. To achieve this, we propose a contrastive learning objective~\cite{oord2018representation} as follows:
\begin{align}
    \label{eq:loss}
    &L(\mathcal{I}, \mathcal{I'}, t, \theta) = \sum_{k=1}^K -\log \frac{\exp{(f[p_{tk}]^\intercal ~f'[p'_{tk}])}}{\sum_{p'} \exp{(f[p_{tk}]^\intercal ~f'[p'])}}\\
    &\text{where~~} f = \mathcal{F}_\theta(\mathcal{I},t), \text{~~~~~~~~~~~~} f' = \mathcal{F}_\theta(\mathcal{I}',t) \nonumber\\
    &\text{~~~~~~~~~~~~} f[p] \text{~is the indexed feature } f \text{~at spatial location ~} p \nonumber
\end{align}

\noindent
here $p_{tk}, p'_{tk}$ are the $k$-th functional keypoints for task $t$ in image images $I,I'$ respectively.
Intuitively, minimizing this objective effectively minimizes the distance between features of functionally corresponding points in the two images (numerator) and maximizes the distance between the feature of a keypoint and the features at all non-corresponding locations $p'$ (denominator). Note that the locations $p'$ includes all keypoint and non-keypoint locations. 

This general contrastive learning formulation can be applied to any convolutional neural network architecture that jointly encodes the image $\mathcal{I}$ and task $t$. In order to model the dependencies between functional keypoints of different tasks, we propose to use a modular architecture allowing us to share filters across tasks. We adopt the architecture proposed in \cite{purushwalkam2019task}. For the sake of completeness, we describe the architecture here in detail.

\subsection{Implementation Details}
Figure~\ref{fig:approach} shows an overview of our proposed model $\mathcal{F}$. For an image $\mathcal{I}$, we first extract task-agnostic features using a ResNet trunk upto the \texttt{conv4\_x} layer (defined in \cite{he2016deep}) as $r = R(\mathcal{I})$. For an image with dimensions $H \times W$, the representation $r$ has spatial dimensions $H/16 \times W/16$ with a $C$ dimensional feature at each location. The representation $r$ is then processed by a modular task-driven feature extractor $\mathcal{T}$ to produce the final features $f = \mathcal{T}(\mathcal{I}, t)$. 

The modular task-driven feature extractor $\mathcal{T}$ consists of $N_T$ layers with each layer comprising of $M_T$ \textit{modules} except for the last layer which comprises of a single module. A module can be any differentiable operation. In our proposed architecture, we use convolution layers with batch normalization\cite{ioffe2015batch} and ReLU activation functions (see supplementary for details of kernel size, number of filters, etc). We denote the $j$-th module of the $i$-th layer as $\mathcal{T}_{ij}$. Given a task-dependent weight tensor $W_t \in \mathbb{R}^{N_T-1 \times M_T \times M_T}$ for task $t$, the output of a module $\mathcal{T}_{ij}$ is computed as:
\begin{align}
    o_{ij} = \mathcal{T}_{ij}\Big(~ \sum_{k=1}^{M_T} W_t[i,j,k] * o_{(i-1)k} ~\Big)
\end{align}

\noindent
Intuitively, the input to a module is a weighted sum of the outputs of the modules in the previous layer. For modules in the first layer, the inputs are taken as the task agnostic representation produced previously \textit{i.e.} $o_{0k} = r$. Finally, the output task-dependent representation is taken as the input of last module $o_{(N_T)1}$.

Note that we assumed that we are given a task-dependent weight tensor $W_t\in \mathbb{R}^{N_T-1 \times M_T \times M_T}$. This weight tensor is estimated using a separate fully-connected neural network $\mathcal{G}$ known as the gating network (see supplementary for parameter details). The gating network takes as input a task-embedding $t$ and outputs the weight tensor as $W_t = \mathcal{G}(t)$. As explained in \cite{purushwalkam2019task}, the input weights to each module ($W[i,j,:]$) needs to be projected to the probability simplex using a softmax operation to encourage separate paths for different tasks. In summary, the output feature representation is computed as $f= \mathcal{T}\big[ R(\mathcal{I}), \mathcal{G}(t) \big]$. 

\subsection{Training}

The objective presented in Equation~\ref{eq:loss} is used to learn the parameters $\theta$ which comprises of the gating network $\mathcal{G}$, modules $\mathcal{T}_{ij}$ and task embeddings $t$ (which are initialized randomly for each task). We pretrain the ResNet model $R$ on ImageNet\cite{deng2009imagenet} and fix its parameters. We optimize the parameters using SGD with a learning rate of 0.01, weight decay of 0.00001 and momentum of 0.9. Each batch consists of 256 pairs of images randomly sampled from the training split of the \dataname~ dataset. 
\section{Experiments}
Modeling functional correspondences provides numerous practical benefits. In this section, we demonstrate this by evaluating our presented model on a suite of tasks. First, we show that our model can effectively identify functional correspondences and outperform numerous baseline methods. We then demonstrate the efficacy of our learned representation for few-shot learning, grasp prediction, and ADROIT manipulation tasks~\cite{Rajeswaran-RSS-18}. 
Note that due to the domain of our training data, we focus our experiments on manipulation related datasets for all tasks. 

\begin{figure*}[h!]
    \centering
    \vspace{-1em}
    \includegraphics[width=0.9\textwidth]{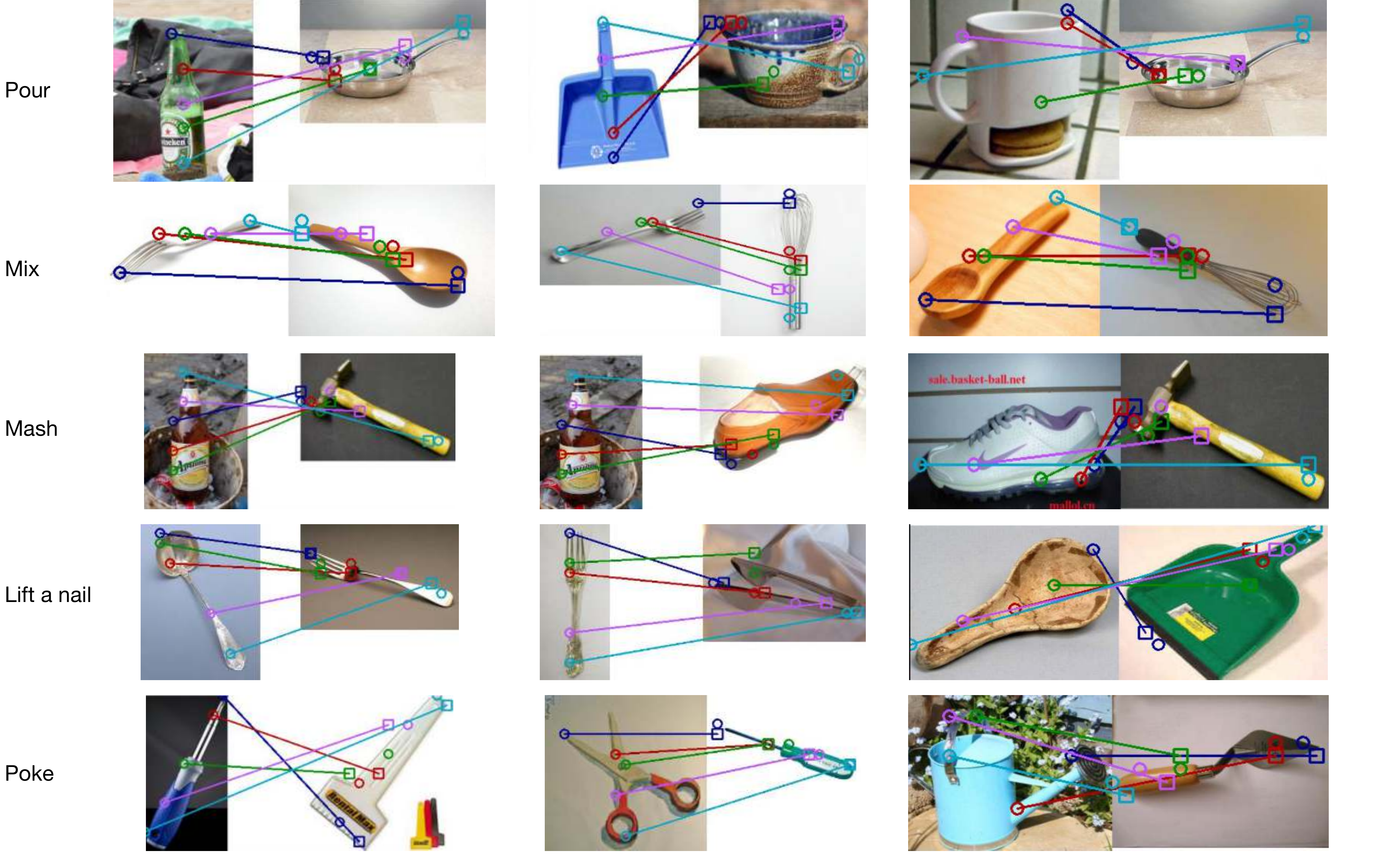}
    \vspace{-1em}
    \caption{Qualitative Correspondences. We demonstrate some qualitative correspondences generated by our algorithm. Squares indicate predicted functionally corresponding keypoints in the second image and circles indicate ground truth keypoints in both images. Notice how the correspondences differ for input task. For example, for the task of mashing/pounding the bottle base corresponds with hammerhead. Similarly for poking, the spout of watering can corresponds to  the tip of gardening tool.}
        \vspace{-1em}
    \label{fig:qualitative}
\end{figure*}

\subsection{Functional Correspondences}
\label{sec:func_corr}
We first evaluate the performance of our model on the task of estimating functional correspondences. We create an evaluation benchmark of (training image,test image, task) triplets using the \dataname dataset. As explained earlier, each image is associated with 5 keypoint annotations. The goal is to identify the location of each keypoint in the test image using the associated train image and task. 

Given representations of the train and test images $f_\text{train}, f_\text{test}$, the corresponding test image location for a training image keypoint $p_\text{train}$ can be identified as:
\begin{align}
 p_\text{test} = \argmax_p (f_\text{train}[p_\text{train}]^\intercal ~f_\text{test}[p])    
\end{align}

We use the PCK metric to evaluate the quality of estimated keypoints. An estimated keypoint in the test image is considered correct if it lies within 23 pixels from the ground truth annotation. 

The ImageNet baseline involves using features from a pretrained ResNet-50 \texttt{conv4\_x} layer (same as the trunk of our model) to compute the correspondences. We also compare to self-supervised semantic correspondence estimation model presented in \cite{Lai20} and features from a task-agnostic affordance estimation method \cite{do2018affordancenet}. Finally, we evaluate three variants of our model. Ours with task-embedding refers to our full model. Ours without task-embedding (uniform) refers to a model with $W_t[i,j,k] = 1/M_T$  \textit{i.e.} the gating weights are constant, uniform and not dependent on the task embedding $t$. Ours without task-embedding (learned) refers to a similar task-independent model, where a single learned gating weight $W_t$ is shared for all tasks. We observe that our proposed model substantially outperforms the ImageNet model and self-supervised learning methods. This further illustrates the difference between learning representations for semantic and functional correspondence. In the ablation of our model, we observe that our proposed model outperforms its task-independent variants by a substantial margin. This emphasizes the need for task-dependent features since functional correspondences are closely tied to the task considered. 

We also train a variant of our task-dependent model by initializing from a self-supervised learning method DINO \cite{caron2021emerging}. We observe that while this underperforms the model initialized from ImageNet, it still significantly outperforms all the baseline methods. This indicates that learning using the FunKPoint dataset is crucial. 

Finally, we measure the consistency of annotations across humans in the functional correspondences by collecting a second set of human annotations. We observe that the new annotations of correspondence achieve a PCK of 82.5\%. Additionally, on a subset of randomly chosen 200 pairs of images, we collected annotations from 4 humans. We observe that the median distance between estimated functional keypoints was 13.07 pixels. These results demonstrate the ambiguity in the functional correspondence task is minimal.

\begin{figure*}[t!]
    \centering
    \vspace{-1em}
    \includegraphics[width=0.9\linewidth]{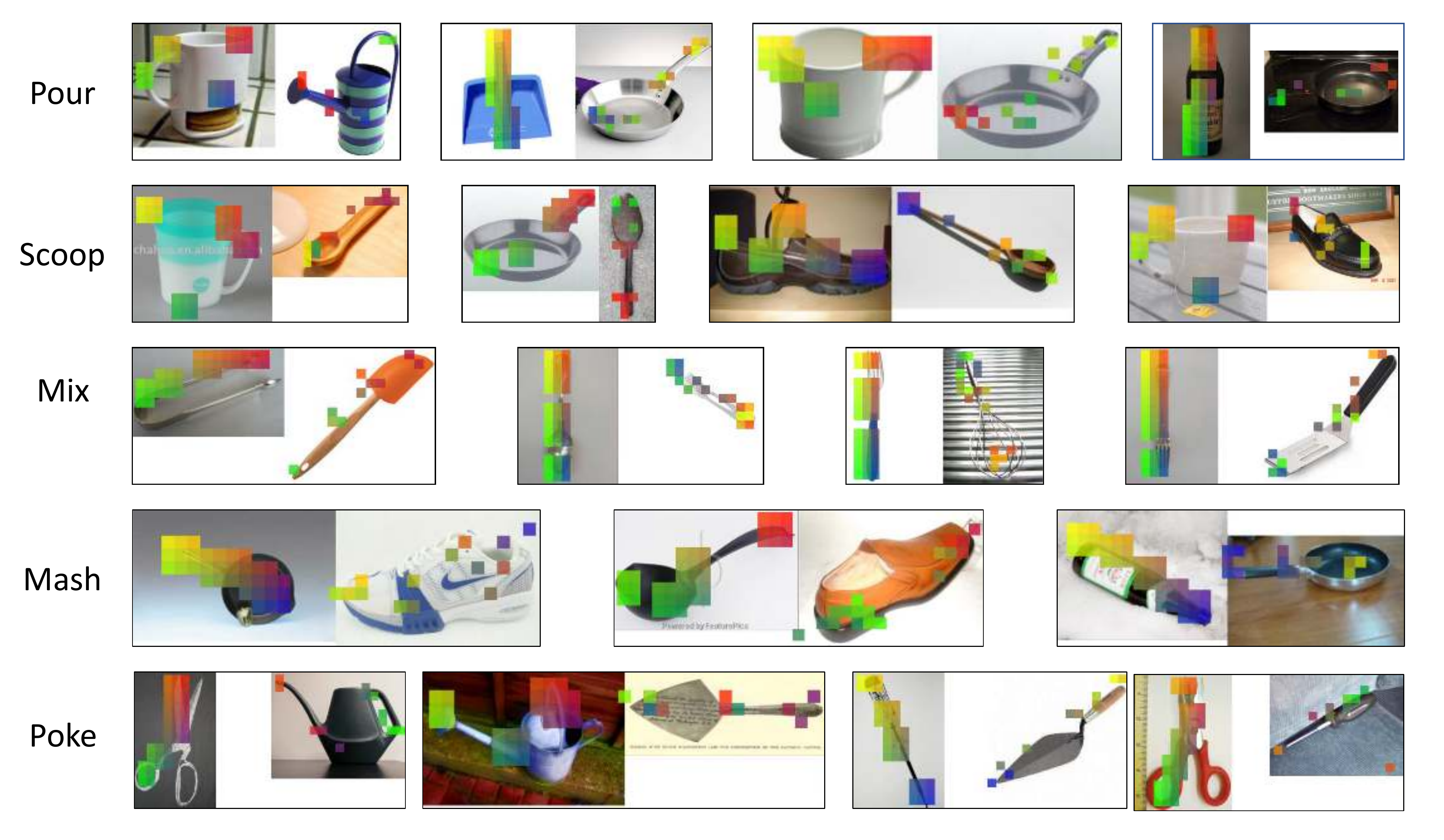}
    \vspace{-1em}
    \caption{\textbf{Beyond Keypoint Correspondences:} The representation learned by our proposed model can be used effectively to identify dense functional correspondences from a reference image (left in each pair) to target images (right in each pair). The colors indicate matched points. Observe that the identified correspondences (same color points) are consistent in terms of functionality.}
    \vspace{-2em}
    \label{fig:densequalitative}
\end{figure*}

In Figure~\ref{fig:qualitative}, we present a visualization of the estimated correspondences for five tasks. We observe that our model is able to learn inter-category correspondences. For example, it is able to learn correspondence between bottlehead and pan spouts for pouring. Some interesting correspondences include correspondence between hammerhead and sole of the shoe and correspondence between spout and tip of gardening-tool. While our model was trained to estimate correspondences for keypoints, our model learns to estimate correspondences for all points on objects. In Figure~\ref{fig:densequalitative}, we visualize densely sampled points on objects and their estimated correspondences on test images. While our model is trained on 5 key points in each image, we observe that the model can approximately associate each densely sampled locations on the reference object to the functionally appropriate location on the target object. For example, the rim of the mug in the first image is appropriately associated with the spout of the watering can. 

\subsection{Few-shot Generalization}
\label{sec:fewshot}

Classification of objects requires understanding its appearance and 3D structure. However, exhaustively modeling appearance and 3D properties from a few samples is challenging and ambiguous in many cases. For example, observing an image of a white conical coffee mug could lead to the belief that all mugs are conical. What we need is a way to use the data from other categories to help learn what makes mug a mug? Since, in the task of functional correspondence, we already label correspondences across multiple categories, our learned model might have better ability to create cross-category generalization. This is the hypothesis we want to test in this experiment. 

\begin{table}[t]
\small
\centering
\caption{\textbf{Correspondences Quantitative Evaluation:}  
}
\vspace{-1em}
\label{tab:pck_seen}
\resizebox{\linewidth}{!}{
\begin{tabular*}{\linewidth}{@{\extracolsep{\fill}}lc@{}}
\toprule
 \textbf{Method}                                & \textbf{PCK} \\ \midrule
 ImageNet (ResNet50)                            & 22.0 \\
 MAST \cite{Lai20} (ResNet18)                        &  8.3 \\
 AffordanceNet \cite{do2018affordancenet} & 15.3 \\
 %UVC \cite{uvc_2019}                                    &       \\
 Ours without task-embedding (uniform)          & 52.8 \\
 Ours without task-embedding (learned)          & 52.5 \\
 Ours with task-embedding                       & 58.4 \\
 Ours with task-embedding (+DINO Init.)         & 43.5 \\
 Human Annotator                                & 82.5 \\
 \bottomrule
\end{tabular*}
}
\vspace{-1em}
\end{table}
\begin{table}[t]
    \small
    \centering
    \caption{\textbf{Fewshot Learning Accuracy: } We observe that the representation learned for functional correspondence (row 3) demonstrates superior generalization in a few-shot learning setup compared to the baseline ImageNet-based representation (row 1) and an ImageNet representation finetuned to classify the objects in the FunKPoint dataset (row 2).  
    }
    \vspace{-1em}
    \label{tab:fewshotclassifier}
    \resizebox{\linewidth}{!}{
    \begin{tabular*}{\linewidth}{@{\extracolsep{\fill}}lccc@{}}
    \toprule
     \multirow{2}{*}{\textbf{Method}} & \multicolumn{3}{c}{\textbf{Accuracy}} \\ 
     \cmidrule(l){2-4}
     & \textbf{1-shot} & \textbf{2-shot} & \textbf{5-shot} \\%& \textbf{NN} \\ 
     \midrule
     ImageNet & 44.68 & 52.52 & 54.63 \\%& 49.00\\
     ImageNet FT FunKPoint & 45.03 & 53.91 & 55.55 \\%& 49.00\\
     Ours & 47.46 & 55.68 & 56.32 \\%& 52.00\\
     \bottomrule
    \end{tabular*}
    }
    \vspace{-2em}
\end{table}
First, we curate a small dataset of 5 manipulable objects (shovel, water jug, coffee mug, wok and letter opener) with 20 images each. We create train-test splits by including 1, 2 or 5 images for each object in the train set and the rest in the test set. In each of the settings, we generate 3 different random samples for the splits leading to a total of 9 unique splits. 

We train a linear classifier to classify the features extracted from our proposed model. Since our model extracts task-dependent features, for each image, we concatenate the features extracted for all 10-tasks and perform spatial average pooling to reduce the dimensionality. As a baseline, we similarly train a linear classifier on the ResNet-50 \texttt{conv4\_x} features pretrained on ImageNet.  For a fair comparison, we also finetune an ImageNet representation to classify objects in our presented FunKPoint dataset. We present the results in Table~\ref{tab:fewshotclassifier}. The representation learned by our model outperforms the ImageNet based representation on all three settings by substantial margins. We also note that our model outperforms the representation optimized for classifying the objects in the FunKPoint dataset. This indicates that the task of functional correspondence leads to representations that generalize better to novel manipulable objects.

\subsection{Grasp Prediction}
\label{sec:grasp}
Functional representations are ideally suited for facilitating downstream robotic manipulation tasks. A common challenge addressed in robotic manipulation is the task of grasp prediction. In ~\cite{pinto2017learning}, this is formalized as prediction of grasp success given an image and a hypothesized grasp angle. For this task, we evaluate the efficacy of the features learned by performing functional correspondence. We extract features using our proposed model, concatenate with the hypothesized grasp angle (as a discretized 18-way one-hot vector), the extracted feature is then fed into a 2-layer neural network appended at the end of the modular network to predict the grasp success label. The feature extractor and the classifier are jointly finetuned on the training set of the benchmark using a smaller learning rate of $2e^{-4}$, until the model converges. 
% We also apply random dropout (with a probability of $0.6$) on the angle input to avoid overfitting. 
% All inputs are scale to a resolution of $256\times 256$ and normalized with the ImageNet statistics. 
As a baseline, we use an ImageNet-based ResNet-50 (similarly truncated at \texttt{layer3} as ours). Table~\ref{tab:grasp_grasp} contains the numerical results for our model on the Grasp benchmark. Our method outperform the baseline method by $1.7\%$ accuracy, demonstrating the advantage of functional representations.

\begin{table}[h]
\small
    \centering
    \caption{\textbf{Classification accuracy on Grasp Dataset~\cite{pinto2017learning}:}  
    }
    \vspace{-1em}
    \label{tab:grasp_grasp}
    % \begin{tabular*}{\linewidth}{@{\extracolsep{\fill}}lc@{}}
    \begin{tabular}{l @{\extracolsep{2cm}} c}
    \toprule
     \textbf{Method} &                                            \textbf{Accuracy} \\ \midrule
     ImageNet & 88.17\\
     ImageNet FT FunKPoint & 88.64 \\
     Ours & \textbf{89.85} \\
     \bottomrule
    \end{tabular}
    \vspace{-1em}
    \end{table}

\subsection{ADROIT Manipulation Task}

\begin{figure}[t]
    \centering
    \includegraphics[width=0.8\linewidth]{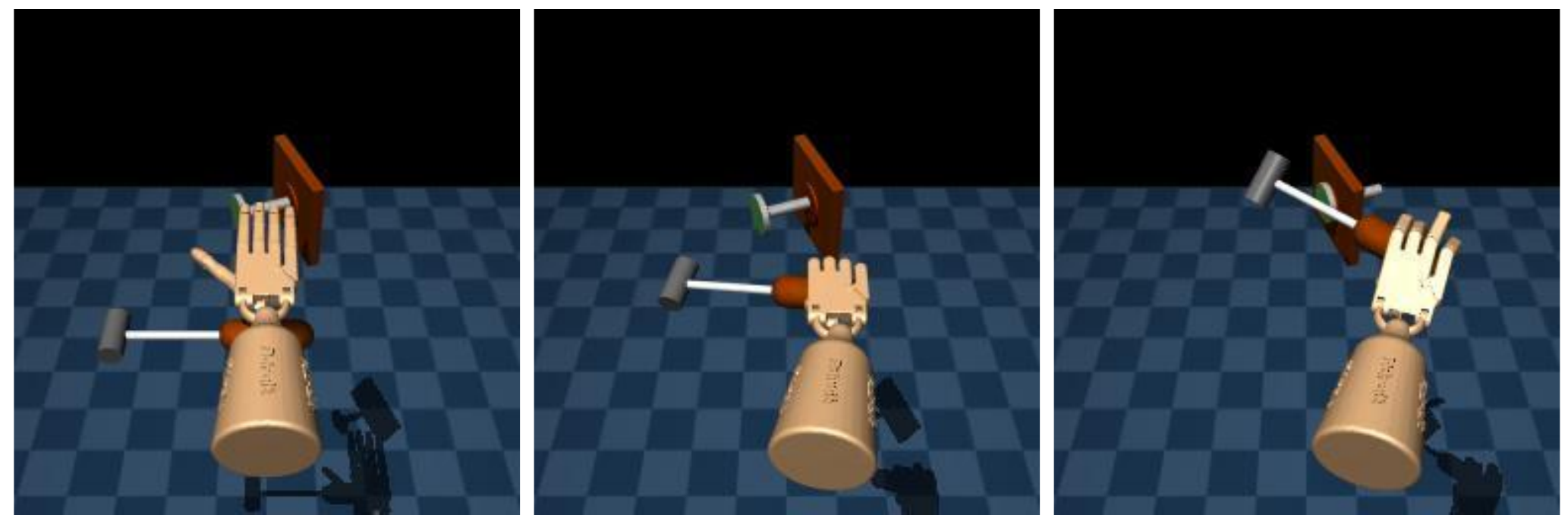}
    % \vspace{-2em}
    \includegraphics[width=0.9\linewidth]{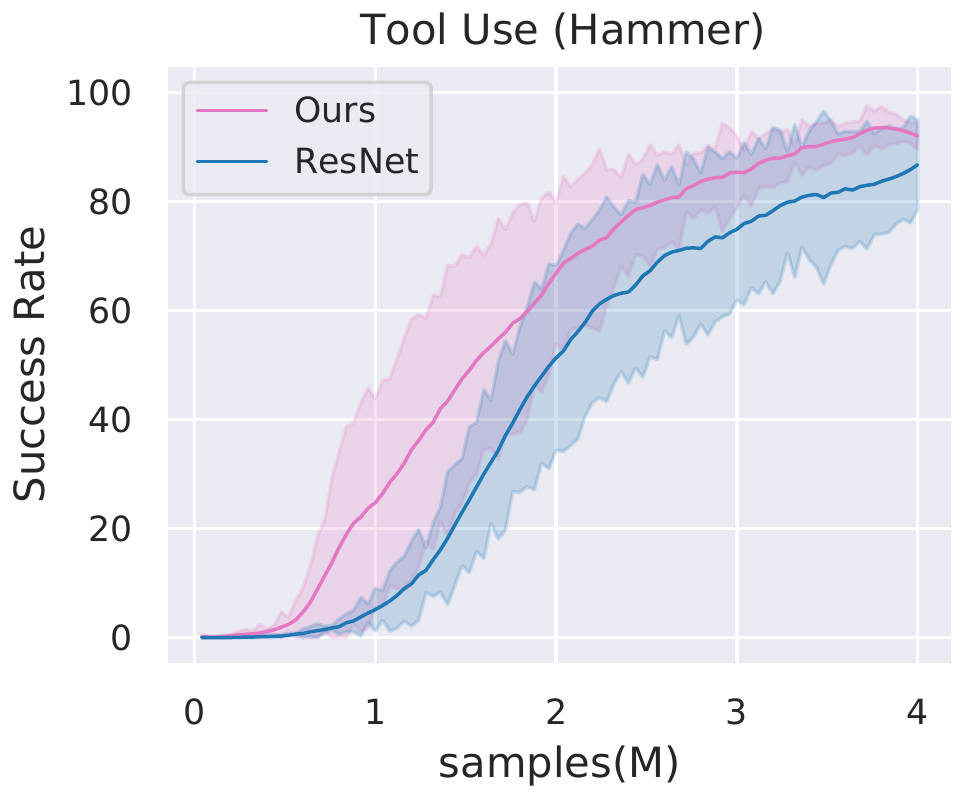}
    \caption{\textbf{RL-based Manipulation:} We evaluate on the hammering a nail Tool Use task (top) introduced in ~\cite{Rajeswaran-RSS-18}. We observe that our functional representation demonstrates improved sample efficiency and success rate compared to a baseline ImageNet-based representations that does not encode functional aspects of objects. }
    % Baselines include NPG (state vector)~\cite{Rajeswaran17Towards}, DAPG (state vector)~\cite{Rajeswaran-RSS-18}, FERM~\cite{zhan2020framework}, and RRL (ImageNet)~\cite{Rutav21RRL}. Note FERM is unstable and NPG, DAPG use state information.}
    \vspace{-2em}
    \label{fig:adroit}
\end{figure}

A lot of recent research has focused on learning robotic manipulation of objects through reinforcement learning. In this section, we investigate whether standard reinforcement learning (RL) based methods can take advantage of functional representations.  
%can benefit from our model: we build an RL algorithm that uses our representation and test it on an RL environment.
We adopt the method proposed in RRL~\cite{Rutav21RRL}, a simple RL algorithm that uses pre-trained ResNet~\cite{he2016deep} features which can be easily replaced by our representation.
We evaluate this algorithm on the ADROIT manipulation suite~\cite{Rajeswaran-RSS-18}, which consists of several complex dexterous manipulation tasks. In the Tool Use task environment, we evaluate for the task of hammering a nail.

In Figure~\ref{fig:adroit}, we present the success rate of our representation compared to the baseline ImageNet-based based features. We observe that our representation leads to improved sample efficiency and final performance at convergence. We believe these results demonstrate the promise of functional representations for robotics problems. We hope that this will inspire more exhaustive investigations of functional representations and their role in robotics. 

\noindent {\bf Acknowledgement:} This research is supported by grants from ONR MURI, the ONR Young Investigator Award to Abhinav Gupta and the DAPRA MCS award.

\appendix
\section{Appendix}
%!TEX root=../main_supp.tex

\subsection{Annotation Interface}
\label{suppsec:interface}
\begin{figure*}[t]
    \centering
    \includegraphics[width=0.95\textwidth]{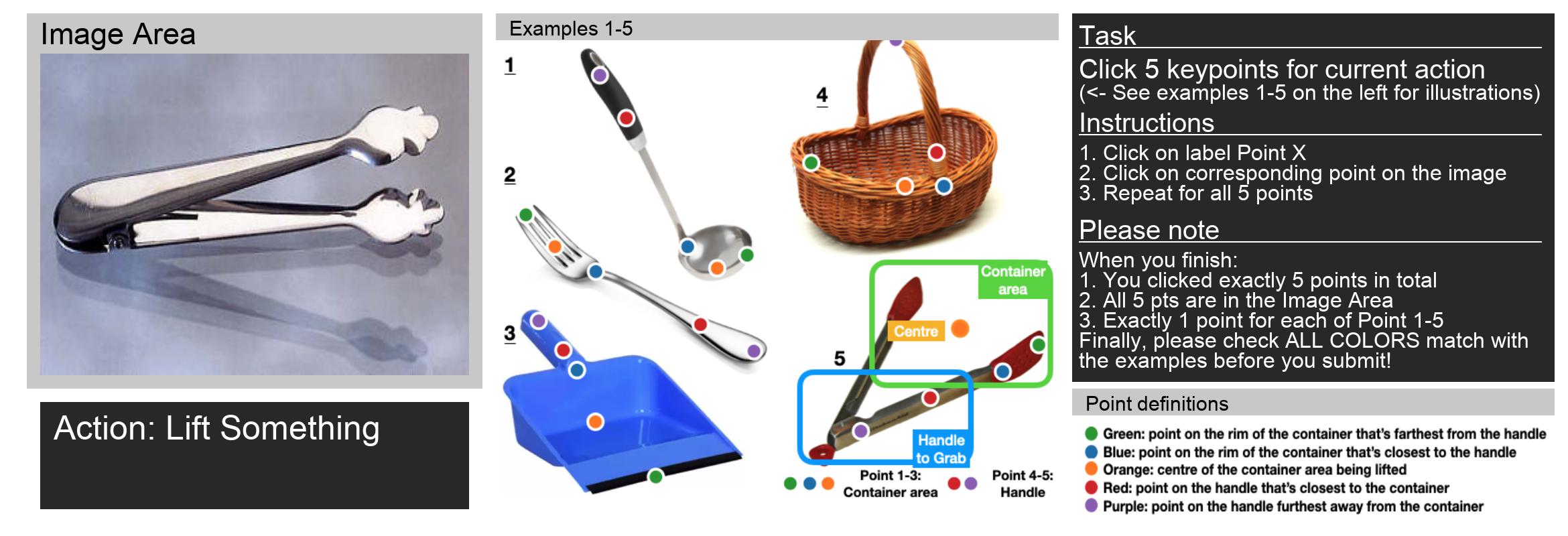}
    % \vspace{-0.3in}
    \caption{\textbf{Annotation interface: } The workers are asked the label the image to the left with instructions and examples given.}
    % \vspace{-0.15in}
    \label{fig:anno}
\end{figure*}

Figure~\ref{fig:anno} shows the annotation interface we used in the Amazon Mechanical Turk system. The image for labelling is shown to the left, together with the specific action we are considering. In the middle, we show 5 examples of labelled images. To the right, we show specific instructions and definitions of each point that is being labelled. Both image examples and point definition are conditioned on the given action. 
Three extra constraints are put on the labelled points:
\begin{enumerate}
    \item The worker must add all keypoint annotations and use each label only once
    \item The worker must annotates all points inside the given ``Image Area''.
    \item The worker must add annotation for the difficulty.
\end{enumerate}

The labelling process took around 5 days. We then check for errors in the annotations and relabel as described in the main text.

\subsection{Annotation Difficulties}
\label{suppsec:difficulty}
\begin{figure}[h]
    \centering
    \includegraphics[width=\columnwidth]{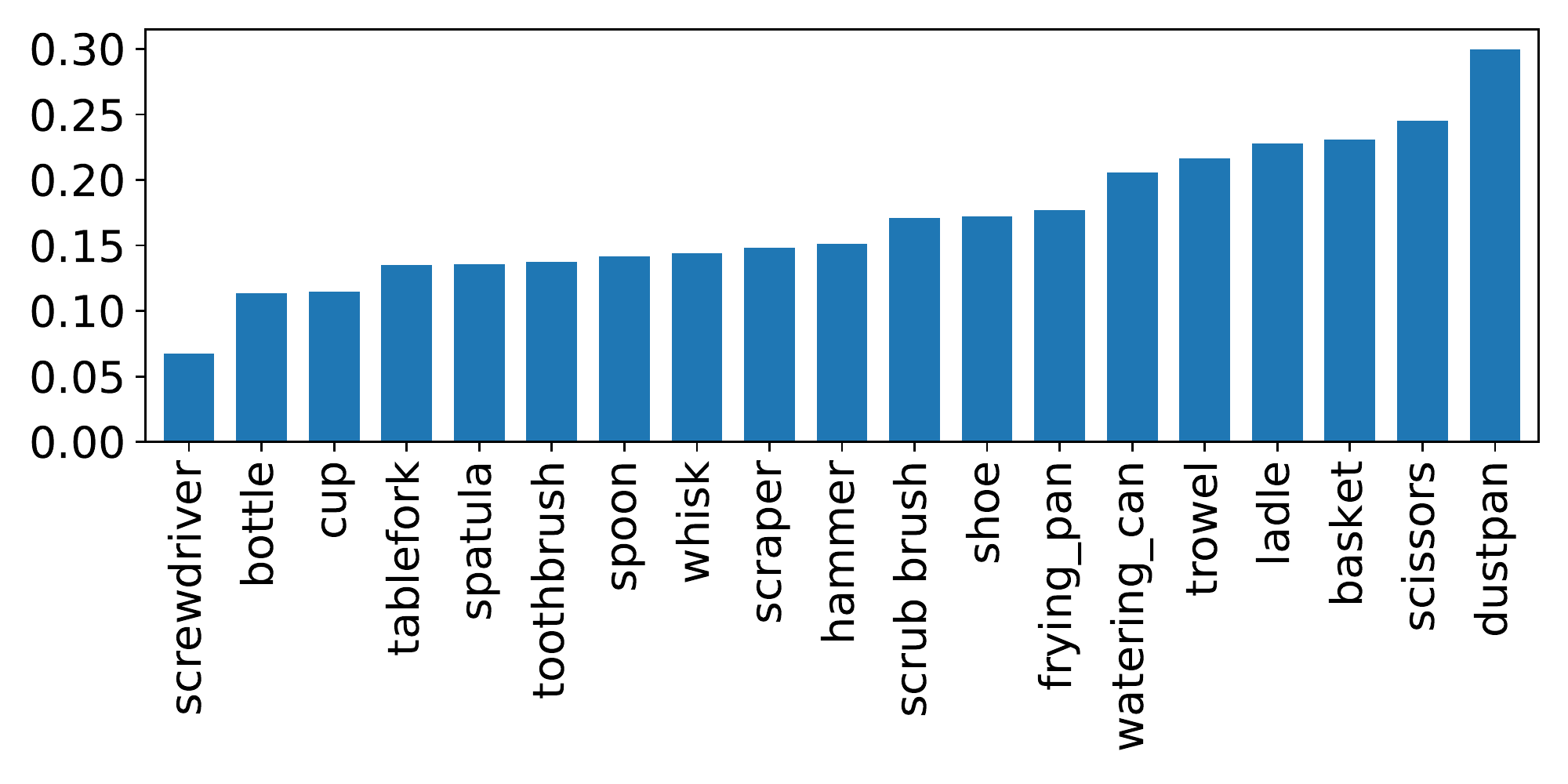}
    \caption{Level of difficulties for each object category.}
    \label{fig:object_diff}
    \vspace{-1em}
\end{figure}

Figure~\ref{fig:object_diff} shows the level of difficulties provided by the annotators. Note annotators could be different for different categories, and the difficulty values may not be consistent across all annotators (different annotators may feel different difficulty for labelling the same image). Here, 0 means \textit{Easy}, 0.5 means \textit{Medium} and 1 means \textit{Hard}. The values shown are computed from an average over all objects in the class. As one could expect, screwdriver is the easiest object category because there are very little ambiguities in defintions of each point and the shape variations are small. Two most difficult object classes are baskets and dustpan. The potential reason could be their large shape variance. For baskets, there are woven basket, shopping basket, basket with lids, without lids, and many others. Similarly, dustpan could have different handle length and orientation. Some so called lobby dustpans could have another structure that functions as a lid. As comparison, the easier object classes such as bottle, cup and tablefork have relatively little shape variance.

\subsection{Implementation Details }
\label{suppsec:hyperparams}

In this section, we provide additional hyperparameter details to facilitate easy reproduction of our results.

As explained in Sec 4.1 of the main text, our proposed model is inspired from the task-driven modular networks proposed in \cite{purushwalkam2019task}. We design the modular network as 4 layers with 6,6,6,1 modules in each layer respectively. Here we present the hyperparameters of the convolution layers:\\
\textbf{1st layer}: 128 filters, kernel size=7, stride=1, padding=3\\
\textbf{2nd layer}: 128 filters, kernel size=3, stride=1, padding=1\\
\textbf{3rd layer}: 128 filters, kernel size=3, stride=1, padding=1\\
\textbf{4th layer}: 128 filters, kernel size=1, stride=1, padding=0

We train the modules using SGD with learning rate 0.01, momentum 0.9 and weight decay 0.00001 with a batch size of 256. The gating network takes as input a 100-dimensional embedding based on the task under consideration. The 10 embeddings for the 10 tasks in the FunKPoint dataset are randomly initialize and learned during the optimization process. The gating network consists of a 2-layer fully-connected neural network with a hidden embedding size of 100.
\section{Dataset Statistics}
As explained in the main text, each object could be associated with multiple actions. This leads to varying number of keypoint annotations based on object category. In Figure \ref{fig:object_count}, we present these statistics:
\begin{figure}[H]
    \centering
    \includegraphics[width=\columnwidth]{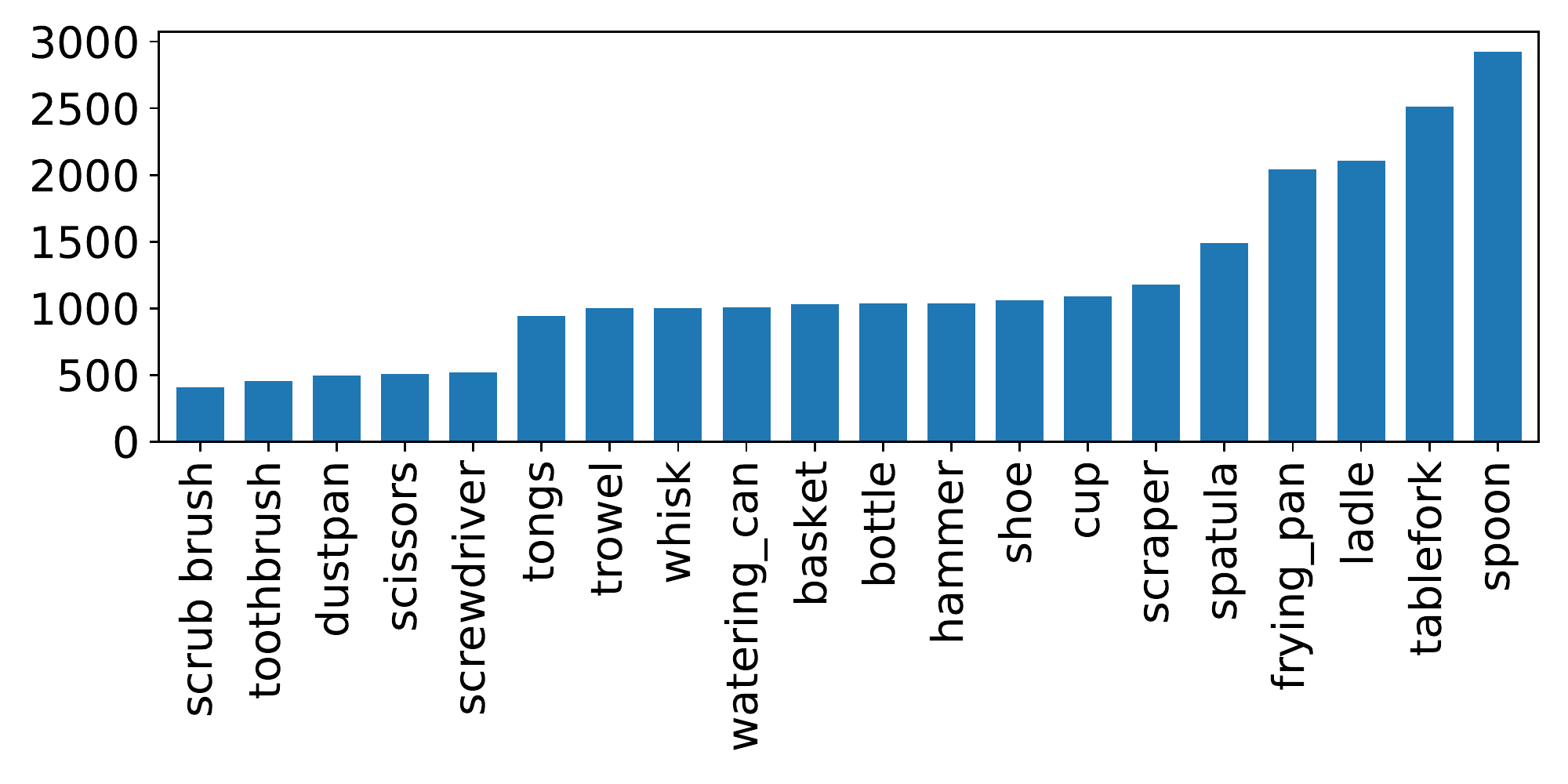}
    \caption{Number of keypoint annotations for each object category.}
    \label{fig:object_count}
    % \vspace{-2em}
\end{figure}

% Push reference to new page
\vspace{10em}
{\small
\bibliographystyle{ieee_fullname}
\bibliography{papers,zihang}

\begin{thebibliography}{10}\itemsep=-1pt

\bibitem{Bailer17}
C. Bailer, K. Varanasi, and D. Stricker.
\newblock Cnn-based patch matching for optical flow with thresholded hinge
  embedding loss.
\newblock 2017.

\bibitem{bakhtin2021residual}
Anton Bakhtin, Yuntian Deng, Sam Gross, Myle Ott, Marc'Aurelio Ranzato, and
  Arthur Szlam.
\newblock Residual energy-based models for text.
\newblock {\em Journal of Machine Learning Research}, 22(40):1--41, 2021.

\bibitem{bay2006surf}
Herbert Bay, Tinne Tuytelaars, and Luc Van~Gool.
\newblock Surf: Speeded up robust features.
\newblock In {\em European conference on computer vision}, pages 404--417.
  Springer, 2006.

\bibitem{berg2005shape}
Alexander~C Berg, Tamara~L Berg, and Jitendra Malik.
\newblock Shape matching and object recognition using low distortion
  correspondences.
\newblock In {\em 2005 IEEE computer society conference on computer vision and
  pattern recognition (CVPR'05)}, volume~1, pages 26--33. IEEE, 2005.

\bibitem{Brox09}
T. Brox, C. Bregler, and J. Malik.
\newblock Large displacement optical flow.
\newblock 2009.

\bibitem{caron2021emerging}
Mathilde Caron, Hugo Touvron, Ishan Misra, Herv{\'e} J{\'e}gou, Julien Mairal,
  Piotr Bojanowski, and Armand Joulin.
\newblock Emerging properties in self-supervised vision transformers.
\newblock {\em arXiv preprint arXiv:2104.14294}, 2021.

\bibitem{Xie13}
S.~Todorovic D.~Xie and S.C. Zhu.
\newblock Inferring ‘dark matter’ and ‘dark energy’ from videos.
\newblock In {\em ICCV}, 2013.

\bibitem{deng2009imagenet}
Jia Deng, Wei Dong, Richard Socher, Li-Jia Li, Kai Li, and Li Fei-Fei.
\newblock Imagenet: A large-scale hierarchical image database.
\newblock In {\em 2009 IEEE conference on computer vision and pattern
  recognition}, pages 248--255. Ieee, 2009.

\bibitem{devin2017learning}
Coline Devin, Abhishek Gupta, Trevor Darrell, Pieter Abbeel, and Sergey Levine.
\newblock Learning modular neural network policies for multi-task and
  multi-robot transfer.
\newblock In {\em 2017 IEEE international conference on robotics and automation
  (ICRA)}, pages 2169--2176. IEEE, 2017.

\bibitem{AffordanceNet18}
Thanh-Toan Do, Anh Nguyen, and Ian Reid.
\newblock Affordancenet: An end-to-end deep learning approach for object
  affordance detection.
\newblock In {\em International Conference on Robotics and Automation (ICRA)},
  2018.

\bibitem{do2018affordancenet}
Thanh-Toan Do, Anh Nguyen, and Ian Reid.
\newblock Affordancenet: An end-to-end deep learning approach for object
  affordance detection.
\newblock In {\em 2018 IEEE international conference on robotics and automation
  (ICRA)}, pages 5882--5889. IEEE, 2018.

\bibitem{du2019implicit}
Yilun Du and Igor Mordatch.
\newblock Implicit generation and modeling with energy based models.
\newblock 2019.

\bibitem{manuelli18}
Peter Florence, Lucas Manuelli, and Russ Tedrake.
\newblock Dense object nets: Learning dense visual object descriptors by and
  for robotic manipulation.
\newblock In {\em CoRL}, 2018.

\bibitem{FouheyDirect}
David Fouhey, Xiaolong Wang, and Abhinav Gupta.
\newblock In defense of the direct perception of affordances.
\newblock In {\em arXiv:1505.01085}, 2015.

\bibitem{Fouhey12}
David~F. Fouhey, Vincent Delaitre, Abhinav Gupta, Alexei~A. Efros, Ivan Laptev,
  and Josef Sivic.
\newblock People watching: Human actions as a cue for single-view geometry.
\newblock In {\em ECCV}, 2012.

\bibitem{Gibson79}
J. Gibson.
\newblock {\em The ecological approach to visual perception}.
\newblock Boston: Houghton Mifflin, 1979.

\bibitem{godard2019digging}
Cl{\'e}ment Godard, Oisin Mac~Aodha, Michael Firman, and Gabriel~J Brostow.
\newblock Digging into self-supervised monocular depth estimation.
\newblock In {\em Proceedings of the IEEE/CVF International Conference on
  Computer Vision}, pages 3828--3838, 2019.

\bibitem{Grabner11}
Helmut Grabner, Juergen Gall, and Luc van Gool.
\newblock What makes a chair a chair?
\newblock In {\em CVPR}, 2011.

\bibitem{grathwohl2019your}
Will Grathwohl, Kuan-Chieh Wang, J{\"o}rn-Henrik Jacobsen, David Duvenaud,
  Mohammad Norouzi, and Kevin Swersky.
\newblock Your classifier is secretly an energy based model and you should
  treat it like one.
\newblock {\em arXiv preprint arXiv:1912.03263}, 2019.

\bibitem{Gupta11}
A. Gupta, S. Satkin, A. Efros, and M. Hebert.
\newblock From {3D} scene geometry to human workspace.
\newblock 2011.

\bibitem{harris1988combined}
Christopher~G Harris, Mike Stephens, et~al.
\newblock A combined corner and edge detector.
\newblock In {\em Alvey vision conference}, volume~15, pages 10--5244.
  Citeseer, 1988.

\bibitem{he2016deep}
Kaiming He, Xiangyu Zhang, Shaoqing Ren, and Jian Sun.
\newblock Deep residual learning for image recognition.
\newblock In {\em Proceedings of the IEEE conference on computer vision and
  pattern recognition}, pages 770--778, 2016.

\bibitem{horn1981determining}
Berthold~KP Horn and Brian~G Schunck.
\newblock Determining optical flow.
\newblock {\em Artificial intelligence}, 17(1-3):185--203, 1981.

\bibitem{Ilg17}
E. Ilg, N. Mayer, T. Saikia, M. Keuper, A. Dosovitskiy, and T. Brox.
\newblock Flownet 2.0: Evolution of optical flow estimation with deep networks.
\newblock 2017.

\bibitem{ioffe2015batch}
Sergey Ioffe and Christian Szegedy.
\newblock Batch normalization: Accelerating deep network training by reducing
  internal covariate shift.
\newblock In {\em International conference on machine learning}, pages
  448--456. PMLR, 2015.

\bibitem{jabri2020space}
Allan Jabri, Andrew Owens, and Alexei~A Efros.
\newblock Space-time correspondence as a contrastive random walk.
\newblock {\em arXiv preprint arXiv:2006.14613}, 2020.

\bibitem{Jia17}
X. Jia, R. Ranftl, and V. Koltun.
\newblock Accurate optical flow via direct cost volume processing.
\newblock 2017.

\bibitem{kendall2017end}
Alex Kendall, Hayk Martirosyan, Saumitro Dasgupta, Peter Henry, Ryan Kennedy,
  Abraham Bachrach, and Adam Bry.
\newblock End-to-end learning of geometry and context for deep stereo
  regression.
\newblock In {\em Proceedings of the IEEE International Conference on Computer
  Vision}, pages 66--75, 2017.

\bibitem{Kim17}
S. Kim, D. Min, B. Ham, S. Jeon, S. Lin, and K. Sohn.
\newblock Fcss: Fully convolutional self-similarity for dense semantic
  correspondence.
\newblock 2017.

\bibitem{krizhevsky2012imagenet}
Alex Krizhevsky, Ilya Sutskever, and Geoffrey~E Hinton.
\newblock Imagenet classification with deep convolutional neural networks.
\newblock {\em Advances in neural information processing systems},
  25:1097--1105, 2012.

\bibitem{Lai20}
Zihang Lai, Erika Lu, and Weidi Xie.
\newblock {MAST}: {A} memory-augmented self-supervised tracker.
\newblock In {\em IEEE Conference on Computer Vision and Pattern Recognition},
  2020.

\bibitem{Lai19}
Zihang Lai and Weidi Xie.
\newblock Self-supervised learning for video correspondence flow.
\newblock 2019.

\bibitem{lecun2006tutorial}
Yann LeCun, Sumit Chopra, Raia Hadsell, M Ranzato, and F Huang.
\newblock A tutorial on energy-based learning.
\newblock {\em Predicting structured data}, 1(0), 2006.

\bibitem{liu2019selflow}
Pengpeng Liu, Michael Lyu, Irwin King, and Jia Xu.
\newblock Selflow: Self-supervised learning of optical flow.
\newblock In {\em Proceedings of the IEEE/CVF Conference on Computer Vision and
  Pattern Recognition}, pages 4571--4580, 2019.

\bibitem{lowe1999object}
David~G Lowe.
\newblock Object recognition from local scale-invariant features.
\newblock In {\em Proceedings of the seventh IEEE international conference on
  computer vision}, volume~2, pages 1150--1157. Ieee, 1999.

\bibitem{lowe2004distinctive}
David~G Lowe.
\newblock Distinctive image features from scale-invariant keypoints.
\newblock {\em International journal of computer vision}, 60(2):91--110, 2004.

\bibitem{manuelli19}
Lucas Manuelli, Wei Gao, Peter Florence, and Russ Tedrake.
\newblock kpam: Keypoint affordances for category level manipulation.
\newblock In {\em ISRR}, 2019.

\bibitem{murali2020taskgrasp}
Adithyavairavan Murali, Weiyu Liu, Kenneth Marino, Sonia Chernova, and Abhinav
  Gupta.
\newblock Same object, different grasps: Data and semantic knowledge for
  task-oriented grasping.
\newblock In {\em Conference on Robot Learning}, 2020.

\bibitem{Novotny17}
D. Novotny, D. Larlus, and A. Vedaldi.
\newblock Anchornet: A weakly supervised network to learn geometry-sensitive
  features for semantic matching.
\newblock 2017.

\bibitem{oord2018representation}
Aaron van~den Oord, Yazhe Li, and Oriol Vinyals.
\newblock Representation learning with contrastive predictive coding.
\newblock {\em arXiv preprint arXiv:1807.03748}, 2018.

\bibitem{pinto2017learning}
Lerrel Pinto and Abhinav Gupta.
\newblock Learning to push by grasping: Using multiple tasks for effective
  learning.
\newblock In {\em 2017 IEEE international conference on robotics and automation
  (ICRA)}, pages 2161--2168. IEEE, 2017.

\bibitem{purushwalkam2019task}
Senthil Purushwalkam, Maximilian Nickel, Abhinav Gupta, and Marc'Aurelio
  Ranzato.
\newblock Task-driven modular networks for zero-shot compositional learning.
\newblock In {\em Proceedings of the IEEE/CVF International Conference on
  Computer Vision}, pages 3593--3602, 2019.

\bibitem{Rajeswaran-RSS-18}
Aravind Rajeswaran, Vikash Kumar, Abhishek Gupta, Giulia Vezzani, John
  Schulman, Emanuel Todorov, and Sergey Levine.
\newblock {Learning Complex Dexterous Manipulation with Deep Reinforcement
  Learning and Demonstrations}.
\newblock In {\em Proceedings of Robotics: Science and Systems (RSS)}, 2018.

\bibitem{Revaud15}
J. Revaud, P. Weinzaepfel, Z. Harchaoui, and C. Schmid.
\newblock Epicflow: Edge-preserving interpolation of correspondences for
  optical flow.
\newblock 2015.

\bibitem{Rivlin}
E. Rivlin, S. Dickinson, and A. Rosenfeld.
\newblock Recognition by functional parts.
\newblock In {\em CVIU}, 1995.

\bibitem{Rocco17}
I. Rocco, R. Arandjelovic, and J. Sivic.
\newblock Convolutional neural network architecture for geometric matching.
\newblock 2017.

\bibitem{Rocco18}
I. Rocco, R. Arandjelovic, and J. Sivic.
\newblock End-to-end weakly-supervised semantic alignment.
\newblock 2018.

\bibitem{Rutav21RRL}
Rutav Shah and Vikash Kumar.
\newblock Rrl: Resnet as representation for reinforcement learning.
\newblock In {\em ICML}, 2021.

\bibitem{simonyan2014very}
Karen Simonyan and Andrew Zisserman.
\newblock Very deep convolutional networks for large-scale image recognition.
\newblock {\em arXiv preprint arXiv:1409.1556}, 2014.

\bibitem{Smith95b}
S.M. Smith and J.M. Brady.
\newblock Asset-2: Real-time motion segmentation and shape tracking.
\newblock {\em IEEE Transactions on Pattern Analysis and Machine Intelligence},
  17(8):814--820, 1995.

\bibitem{Stark}
L. Stark and K. Bowyer.
\newblock Achieving generalized object recognition through reasoning about
  association of function to structure.
\newblock In {\em PAMI}, 1991.

\bibitem{ufer2017deep}
Nikolai Ufer and Bjorn Ommer.
\newblock Deep semantic feature matching.
\newblock In {\em Proceedings of the IEEE conference on computer vision and
  pattern recognition}, pages 6914--6923, 2017.

\bibitem{Wang21}
Jiashun Wang, Huazhe Xu, Jingwei Xu, Sifei Lu, and Xiaolong Wang.
\newblock Synthesizing long-term 3d human motion and interaction in 3d scenes.
\newblock In {\em CVPR}, 2021.

\bibitem{BingeWatching}
Xiaolong Wang, Rohit Girdhar, and Abhinav Gupta.
\newblock Binge watching: Scaling affordance learning from sitcoms.
\newblock In {\em arXiv:1804.03080}, 2018.

\bibitem{Wang19}
X. Wang, A. Jabri, and A. Efros.
\newblock Learning correspondence from the cycle-consistency of time.
\newblock 2019.

\bibitem{wang2019tafe}
Xin Wang, Fisher Yu, Ruth Wang, Trevor Darrell, and Joseph~E Gonzalez.
\newblock Tafe-net: Task-aware feature embeddings for low shot learning.
\newblock In {\em Proceedings of the IEEE/CVF Conference on Computer Vision and
  Pattern Recognition}, pages 1831--1840, 2019.

\bibitem{Binford}
P. Winston, T. Binford, B. Katz, and M. Lowry.
\newblock Learning physical description from functional definitions, examples
  and precedents.
\newblock In {\em MIT Press}, 1984.

\bibitem{yang2020multi}
Ruihan Yang, Huazhe Xu, Yi Wu, and Xiaolong Wang.
\newblock Multi-task reinforcement learning with soft modularization.
\newblock {\em arXiv preprint arXiv:2003.13661}, 2020.

\bibitem{Zhao13}
Y. Zhao and S.C Zhu.
\newblock Scene parsing by integrating function, geometry and appearance
  models.
\newblock 2013.

\end{thebibliography}
}

\end{document}